\pdfoutput=1

\documentclass[11pt]{article}

\usepackage[preprint]{acl}

\usepackage{times}
\usepackage{latexsym}

\usepackage[T1]{fontenc}

\usepackage[utf8]{inputenc}

\usepackage{microtype}

\usepackage{inconsolata}

\usepackage{amsmath,amsfonts,bm}

\def\eqref#1{equation~\ref{#1}}

\def\1{\bm{1}}

\DeclareMathAlphabet{\mathsfit}{\encodingdefault}{\sfdefault}{m}{sl}
\SetMathAlphabet{\mathsfit}{bold}{\encodingdefault}{\sfdefault}{bx}{n}

\DeclareMathOperator*{\argmax}{arg\,max}

\usepackage{times}
\usepackage{latexsym}

\usepackage{hyperref}
\usepackage{url}
\usepackage{booktabs}       
\usepackage{amsfonts}       
\usepackage{subcaption}
\usepackage{graphicx}
\usepackage{amsmath}
\usepackage{mathtools}
\usepackage{tcolorbox}
\usepackage{multirow}

\usepackage{algpseudocode,algorithm,algorithmicx}

\algrenewcommand\algorithmicrequire{\textbf{Input:}}

\newcommand{\sysname}{\texttt{PASS}}
\newcommand \LtlUntil      {\mathbin{\mathcal{U}\kern-.1em}}
\newcommand \LtlNext      {\mathbin{\bigcirc\kern.1em}}
\newcommand \LtlEventually     {\mathbin{\lozenge\kern.1em}}
\newcommand \LtlAlways      {\mathbin{\square\kern.1em}}

\title{Formally Specifying the High-Level Behavior of LLM-Based Agents}

\author{Maxwell Crouse, Ibrahim Abdelaziz, Ramon Astudillo, Kinjal Basu, Soham Dan, \\ 
        {\bf Sadhana Kumaravel, Achille Fokoue, Pavan Kapanipathi, Salim Roukos, Luis Lastras } \\ 
        {\texttt{ \{ maxwell.crouse, ibrahim.abdelaziz1, ramon.astudillo \}@ibm.com }} \\
        {\texttt{ \{ kinjal.basu, soham.dan, sadhana.kumaravel1 \}@ibm.com }} \\
        {\texttt{ \{ achille, kapanipa, roukos, lastrasl \}@us.ibm.com }}
        \\ IBM Research}

\begin{document}
\maketitle
\begin{abstract}
Autonomous, goal-driven agents powered by LLMs have recently emerged as promising tools for solving challenging problems without the need for task-specific finetuned models that can be expensive to procure. Currently, the design and implementation of such agents is ad hoc, as the wide variety of tasks that LLM-based agents may be applied to naturally means there can be no one-size-fits-all approach to agent design. In this work we aim to alleviate the difficulty of designing and implementing new agents by proposing a minimalistic generation framework that simplifies the process of building agents. The framework we introduce allows the user to define desired agent behaviors in a high-level, declarative specification that is then used to construct a decoding monitor which guarantees the LLM will produce an output exhibiting the desired behavior. Our declarative approach, in which the behavior is described without concern for how it should be implemented or enforced, enables rapid design, implementation, and experimentation with different LLM-based agents. We demonstrate how the proposed framework can be used to implement recent LLM-based agents (e.g., ReACT), and show how the flexibility of our approach can be leveraged to define a new agent with more complex behavior, the Plan-Act-Summarize-Solve (\sysname{}) agent. Lastly, we demonstrate that our method outperforms other agents on multiple popular reasoning-centric question-answering benchmarks.


\end{abstract}

\section{Introduction}


Many recent works (e.g., \cite{brohan2023can,shen2023hugginggpt,yao2022react,shinn2023reflexion}) have explored the use of large language models (LLMs) to drive the decision-making of intelligent, autonomous agents. Given a problem to solve, these LLM-based agents break the problem down into a sequence of steps, where each step involves either generating text or executing a tool, e.g., an API call \cite{schick2023toolformer,qin2023toolllm,tang2023toolalpaca}, which can supply new context to the agent. Importantly, while the order of steps to take is dictated by a high-level, prespecified behavior implemented by the user, the underlying LLM is still allowed significant flexibility in what it may produce. At each individual step, the outputs are entirely determined by the LLM, thus allowing the agent to leverage the strong generative capabilities of LLMs while ensuring there are guardrails to prevent aberrant behavior. Figure \ref{fig:react_text} provides an example of text produced by the popular ReACT \cite{yao2022react} framework, where an agent executes a loop that goes between generative steps (i.e., \texttt{Thought}, \texttt{Action}, \texttt{Action Input}) and tool execution steps (i.e., \texttt{Observation}).

\begin{figure}[t]
\begin{center}
\begin{tcolorbox}[width=0.97\linewidth,colback=white!90!black]
{\small
\texttt{\textbf{[Question]} Musician and satirist Allie Goertz wrote a song about the ``The Simpsons'' character Milhouse, who Matt Groening named after who? \\
\textbf{[Thought]} The question simplifies to ``The Simpsons'' character Milhouse is named after who. I only need to search ... \\
\textbf{[Action]} Search \\
\textbf{[Action Input]} Milhouse \\
\textbf{[Observation]} Milhouse Mussolini Van Houten is a recurring character in ... \\ 
\textbf{[Thought]} The paragraph does not tell who Milhouse is named after, maybe I can look up ``named after''. \\
\textbf{[Action]} Lookup \\
\textbf{[Action Input]} named after \\
\textbf{[Observation]} (Result 1 / 1) Milhouse was named after U.S. president Richard Nixon, whose middle name was Milhous. \\
\textbf{[Final Thought]} Milhouse was named after U.S. president Richard Nixon, so the answer is Richard Nixon. \\
\textbf{[Answer]} Richard Nixon
}
}
\end{tcolorbox}
\caption{Text output by ReACT agent that alternates between generative states (\texttt{Thought}, \texttt{Action}, \texttt{Action~Input}) and tool execution states (\texttt{Observation})}
\label{fig:react_text}
\end{center}
\end{figure}


Though LLM-based agents show much promise, there still remain challenges involved with their practical application. As each agent has its own strengths and weaknesses, it can be necessary to try a variety of different agents when approaching a problem. This can be a steep barrier to entry, as the lack of a standard framework for defining agents means that the end user must reimplement in code the exact behavior they wish for an agent to exhibit. In addition, the use of explicit code to implement agents has lead to their execution being largely rigid, i.e., they are hard coded to follow a fixed path of behavior; which takes away flexibility from the LLM in deciding how best to solve a problem. 

To address the aforementioned challenges, we propose a declarative framework to formally specify the high-level behavior of an LLM-based agent. Our framework takes in an agent behavior specified as a finite-state machine, which it then uses to define a decoding monitor that ensures the LLM-based agent executes steps in a way that conforms to the user's expectation. 
Importantly, the decoding monitor operates in a post hoc fashion, intervening only to correct generated text when it observes a deviation from the desired behavior. This makes it applicable to models only accessible through APIs. 

The ability to leverage the highest performing LLMs (which are often closed source) is a key component of our approach. Following in-context learning \cite{brown2020language} and instruction-tuning \cite{wei2021finetuned,ouyang2022training}, LLMs have shown generalization to a large number of tasks without any parameter tuning. In addition, it is clear that large model size and the use of specialized hardware (e.g., GPUs) are fundamental factors in performance. For these reasons, centralized systems that serve large numbers of requests remotely through an API are increasingly central to how LLMs are utilized. In this context, custom applications of token-level constrained decoding \cite{wuebker-etal-2016-models, hokamp2017lexically} that directly modify the output softmax of a model would have high communication overhead. The decoding monitor proposed here circumvents this by optimistically assuming most generated tokens will be correct and rejects or prefixes the generation of entire chunks of text at the client side, thus reducing communication overhead. 

We demonstrate how a number of popular agents can be straightforwardly implemented in our framework (e.g., ReACT \cite{yao2022react}, ReWOO \cite{xu2023rewoo}, Reflexion \cite{shinn2023reflexion}). In addition, we introduce the Plan-Act-Summarize-Solve (\sysname{}) agent. The \sysname{} agent leverages the flexibility enabled by our framework and operates by dynamically adjusting the number of actions it executes in parallel. It thus differs from prior agents that operate entirely sequentially (e.g., \cite{yao2022react}) or in parallel (e.g., \cite{xu2023rewoo}).




In summary, our contributions in this work are as follows: (a) we introduce a declarative framework for defining LLM-based agents that ensures conformance to desired behaviors with a decoding montitor, (b) we demonstrate how to implement a number of well-known agents with our framework, and (c) we introduce \sysname{}, a new agent architecture that leverages the declarative nature of our framework and yields improved performance as compared to other agents across three standard datasets (Hotpot QA \cite{yang2018hotpotqa}, TriviaQA \cite{joshi-etal-2017-triviaqa}, and GSM8K \cite{cobbe2021training}).



\section{Agent Specification Framework}

In this section, we introduce our framework for designing and implementing autonomous agents that can interact with the environment to solve problems expressed in natural language. Our framework is intended to be lightweight (i.e., add as little additional overhead to LLM operation as is possible) and declarative (i.e., the user specifies the desired high-level behavior in terms of constraints without concern for how they should be implemented or enforced). To begin, we provide a more formal definition of agents and how they are specified in our framework. Then, we describe how the framework is used to control what an agent can generate.



\begin{figure}[t]
\begin{center}
\includegraphics[width=\linewidth]{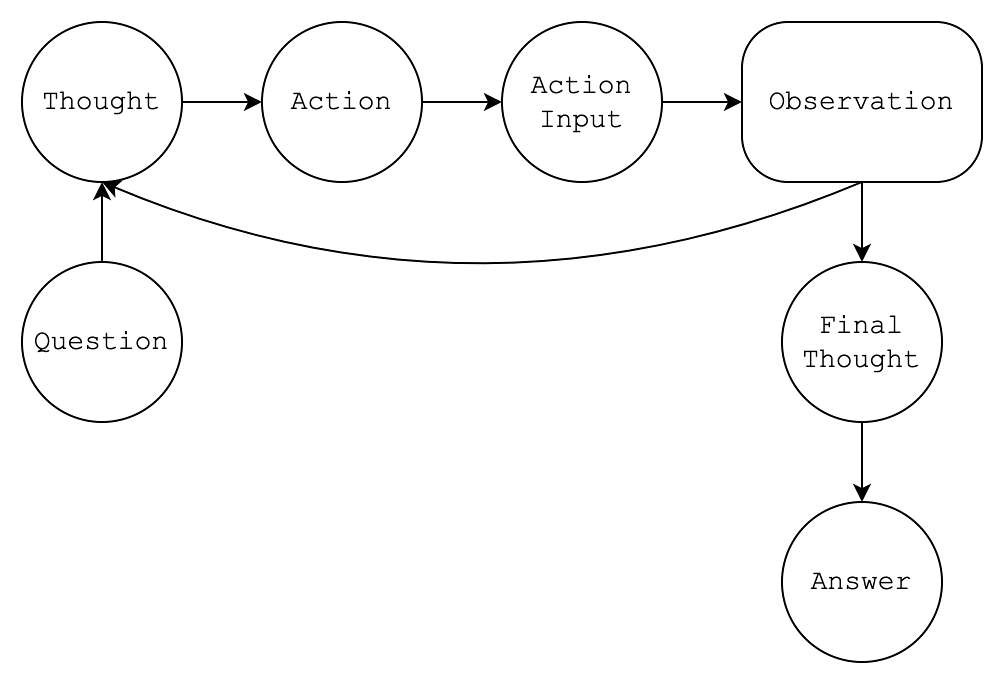}
\caption{State diagram of ReACT agent architecture}
\label{fig:react_arch_diagram}
\end{center}
\end{figure}

\subsection{Specifying Agent Behavior}

We model agents as generic finite-state machines, where a finite-state machine is considered a tuple $\left< \mathcal{D}_S, \delta, s_{0}, s_{end} \right>$ consisting of a non-empty set of states $\mathcal{D}_S$, a state transition function $\delta : \mathcal{D}_S \rightarrow \mathcal{D}_S$, an initial state $s_{0} \in \mathcal{D}_S$, and a final state $s_{end} \in \mathcal{D}_S$.  To define an agent and its underlying state machine, the user provides a specification consisting of 1) a list of states and their properties and 2) a desired behavior in the form of a logical formula. At each time step, the agent will receive a string from either an LLM or the environment\footnote{Here we refer to the ``environment'' as any provider of text that is \textit{not} the LLM (e.g., external tools, API calls, etc.)}, with the source of the string determined by the state the agent is in.

Figure \ref{fig:react_spec} shows an example of a specification provided in the format of a Lisp-style s-expression. In the specification, the \texttt{:states} list contains all possible states for the agent. Each state within the list must specify a prompt string (e.g., ``[Thought]'' for the \texttt{Tht} state), which will serve as both an initial prompt for when the agent is in that state and as a signal to detect when a state transition occurs. When the environment is the intended provider of a string for a particular state, the special \texttt{:env-input} flag is used (e.g., shown in the \texttt{Obs} state).

\begin{figure}[t]
\begin{center}
\begin{tcolorbox}[width=0.97\columnwidth,colback=white!90!black]
{\small
\texttt{(define react-agent \\
\null\quad(:states \\
\null\quad\quad(Ques (:text "[Question]")) \\
\null\quad\quad(Tht (:text "[Thought]")) \\
\null\quad\quad(Act (:text "[Action]")) \\
\null\quad\quad(Act-Inp (:text "[Action Input]")) \\
\null\quad\quad(Obs (:text "[Observation]") \\
\null\quad\quad\quad\quad \ (:flags :env-input)) \\
\null\quad\quad(Final-Tht (:text "[Final Thought]")) \\
\null\quad\quad(Ans (:text "[Answer]"))) \\
\null\quad \\
\null\quad(:behavior \\
\null\quad\quad(next \\
\null\quad\quad\quad Ques \\
\null\quad\quad\quad(until \\
\null\quad\quad\quad\quad(next Tht Act Act-Inp Obs) \\
\null\quad\quad\quad\quad Final-Tht) \\
\null\quad\quad\quad Ans)))
}
}
\end{tcolorbox}
\end{center}
\caption{Specification for ReACT agent (see Figure~\ref{fig:react_arch_diagram}) }
\label{fig:react_spec}
\end{figure}

The behavior of an agent is provided in the \texttt{:behavior} list as a logical formula, where each formula is constructed from states connected together with the logical operators \texttt{or}, \texttt{next} or \texttt{until}. We enforce that the top-level formula in \texttt{:behavior} begins with \texttt{next}. From the formula, a finite-state machine (FSM) is constructed that will be used to validate the behavior of the agent.

The operators are taken from linear-temporal logic (LTL) \cite{pnueli1977temporal}, which is commonly used for runtime verification systems. We note however, that any method for constructing an FSM would suffice. We next give an overview of how our approach evaluates formulas, with a more detailed background of LTL provided in Appendix \ref{sec:ltl_overview}.

In this work, the formula is evaluated over a finite sequence of states, where each state is taken from the list of states provided in \texttt{:states}. In the following, the letters after the logical operator (e.g., \texttt{a}, \texttt{b}, and \texttt{c}) will denote arguments, i.e., formulas or states, and let $\texttt{S} = \left< \texttt{s}_0, \texttt{s}_{1}, \ldots \right>$ denote the sequence of states produced by our agent, with $\texttt{s}_i$ being the state of our agent at time $i$. We write that $\texttt{S}$ satisfies ($\models$) the behavior if the following holds
\begin{alignat*}{4}
&\texttt{S} \ \ &&\models \ \ \texttt{a} \\
& && \ \ \mathrm{iff} \ \ \texttt{S} = \left<\texttt{a}\right> \wedge \texttt{a} \in \texttt{:states} \\
&\texttt{S} \ \ &&\models \ \ \texttt{(or a b c }\ldots\texttt{)} \\
& && \ \ \mathrm{iff} \ \ \texttt{S} \models \texttt{a} \vee \texttt{S} \models \texttt{b} \vee \ldots \vee \texttt{S} \models \texttt{c} \vee \ldots \\
&\texttt{S} \ \ &&\models \ \ \texttt{(next a b c }\ldots\texttt{)} \\
& && \ \ \mathrm{iff} \ \ \exists i > 0 . \ \ \texttt{S}[0 \ldots i] \models \texttt{a} \\
& && \ \ \mathrm{and} \ \ \texttt{S}[i\ldots] \models \texttt{(next b c }\ldots\texttt{)} \\
&\texttt{S} \ \ &&\models \ \ \texttt{(until a b)} \\
& && \ \ \mathrm{iff} \ \ \exists j \geq 0 . \ \ \texttt{S}[j\ldots] \models \texttt{b} \\
& && \ \ \mathrm{and} \ \ \texttt{S}[i \ldots] \models \texttt{a}, \ \ \textrm{for all} \ \ 0 \leq i < j
\end{alignat*}
where $\texttt{S}[i \ldots j] = \left< \texttt{s}_i, \ldots, \texttt{s}_{j-1} \right>$ and $\texttt{S}[i\ldots] = \left< \texttt{s}_i, \ldots \right>$ are slicing operations that return subsequences of observations following time step $i$. 

Informally, the above rules operate as one might expect. \texttt{(next a b c }\ldots\texttt{)} specifies that \texttt{a} must hold, then \texttt{b}, then \texttt{c}, etc. Similarly, \texttt{(until a b)} specifies that \texttt{a} must hold (and may loop indefinitely) until \texttt{b} holds. Lastly, \texttt{(or a b c }\ldots\texttt{)} requires that any of \texttt{a}, \texttt{b}, \texttt{c}, etc. hold. Despite the simplicity of the above representation, we find it to be sufficient to represent a range of agents. We refer back to Figures \ref{fig:react_arch_diagram} and \ref{fig:react_spec} for ReACT, with examples of other agents provided in Appendix \ref{sec:ltl_overview}.



\begin{figure}[t]
\begin{center}
\includegraphics[width=\linewidth]{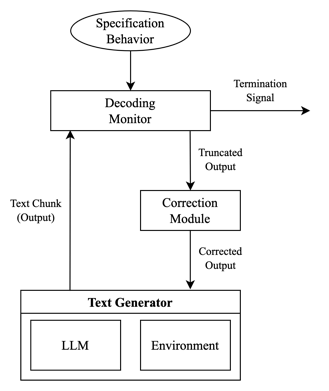}
\caption{Agent generation loop}
\label{fig:sys_design}
\end{center}
\end{figure}

\subsection{Constraining Agent Behavior}


Agents in our framework begin generation like any other prompting approach. First, the agent is provided with the input prompt, which consists of instructions, (optionally) a number of examples, and the prompt text $t_0$ associated with the initial state $s_0$ (e.g., ``[Question]'' for the agent of Figure \ref{fig:react_spec}). At this point, the agent is considered to be in state $s_0$ and must transition between states according to the behavior prescribed by its logical formula. For the remainder of generation, the agent operates in a loop that alternates between generation, validation, and correction, until the the agent reaches the final state $s_{end}$, at which point a termination signal halts generation. Figure \ref{fig:sys_design} provides a diagram illustrating our system's process, which we next describe.

\subsubsection{Generation}

Depending on the current state of the agent, text will be produced by either the underlying LLM of the agent or by the environment (e.g., the result of an API call). When text must be produced by the LLM, the agent's LLM is prompted with all historical context concatenated with all text produced thus far. It then generates text until either a stop sequence is output or a pre-specified chunk size is reached. In this work, stop sequences are the prompt texts for any state that must be produced by the environment (e.g., ``[Observation]'' in Figure \ref{fig:react_spec}). 
When text is produced by the environment, it is treated in the same way as if it were produced by the LLM. In both cases, the chunk of text is then passed to decoding monitor for validation.


\subsubsection{Validation} 

Text received from generation is first parsed and separated into a sequence of states paired with their corresponding content, i.e., $\texttt{S} = \left< \left<s_i, t_i\right>, \ldots \right>$, with separation dividing the text by occurrences of the prompts associated with \textit{any} of the agent's specification's states (regardless of whether the state reflects a valid transition). Splitting on prompt strings thus serves as the state transition function $\delta$, where one must be sure that the prompts for each state are not otherwise commonly occurring strings.

With $\texttt{S}$ in hand, the decoding monitor (constructed from the specification's behavior) walks through each pair $\left< s_i, t_i \right>$ to both update its current state as well as validate the correctness of state transitions. When it detects a state transition error, i.e., when $s_i$ does not follow from $s_{i-1}$ according to the specification's behavior or $s_i$ is not given, the text following the invalid state or state content is discarded, and the remaining subsequence of $S$ is passed to the correction module.


\subsubsection{Correction}


The correction module will be passed the sequence $\texttt{S} = \left< \left<s_i, t_i\right>, \ldots \right>$ from before that has been truncated to remove all text beyond the first violation of the specification. It takes the truncated text and applies a correction, then returns the new string to the LLM to resume generation. The difficulty with this step is to apply a correction that meaningfully changes what the LLM generates next while also not imposing a hard constraint on the LLM.

To correct a state transition error (e.g., the LLM produced ``[Thought]'' instead of ``[Action]''), the correction module employs \textit{valid state prefixing}. Taking the last observed state in $S$ as the last correct state, we define the set of next valid states to be the states that can be transitioned to according to the agent's behavior. Valid state prefixing takes the longest common prefix of the prompt texts for each state within the set of next valid states and appends that prefix to the string to be returned to the LLM. For instance, for ``[Action]'' and ``[Action Input]'', the longest common prefix would be ``[Action'', while for ``[Action]'' and ``[Thought]'', it would be ``[''. Once the longest common prefix is determined, the correction module returns the truncated original text (i.e., all text occurring before the detected error) concatenated with this prefix. 



\section{The \sysname{} Agent Architecture}

\label{sec:pass}

\begin{figure}[t]
\begin{center}
\begin{tcolorbox}[width=0.97\columnwidth,colback=white!90!black]
{\small
\texttt{(define pass-agent \\
\null\quad(:states \\
\null\quad\quad(Ques (:text "[Question]")) \\
\null\quad\quad(Plan (:text "[Thought]")) \\
\null\quad\quad(Act (:text "[Action]")) \\
\null\quad\quad(Act-Inp (:text "[Action Input]")) \\
\null\quad\quad(Sum (:text "[Summary]") \\
\null\quad\quad\quad\quad \ (:flags :env-input)) \\
\null\quad\quad(Final-Tht (:text "[Final Thought]")) \\
\null\quad\quad(Ans (:text "[Answer]"))) \\
\null\quad \\
\null\quad(:behavior \\
\null\quad\quad(next \\
\null\quad\quad\quad Ques \\
\null\quad\quad\quad(until \\
\null\quad\quad\quad\quad(next \\
\null\quad\quad\quad\quad\quad Plan \\
\null\quad\quad\quad\quad\quad(until \\
\null\quad\quad\quad\quad\quad\quad (next Act Act-Inp) \\
\null\quad\quad\quad\quad\quad\quad Sum)) \\
\null\quad\quad\quad\quad Final-Tht) \\
\null\quad\quad\quad Ans)))
}
}
\end{tcolorbox}
\end{center}
\caption{Specification for \sysname{} agent that that iteratively aggregates sets of actions, executes them in parallel, and then summarizes their output}
\label{fig:pass_spec}
\end{figure}

\begin{figure}[t]
\begin{center}
\includegraphics[width=\columnwidth]{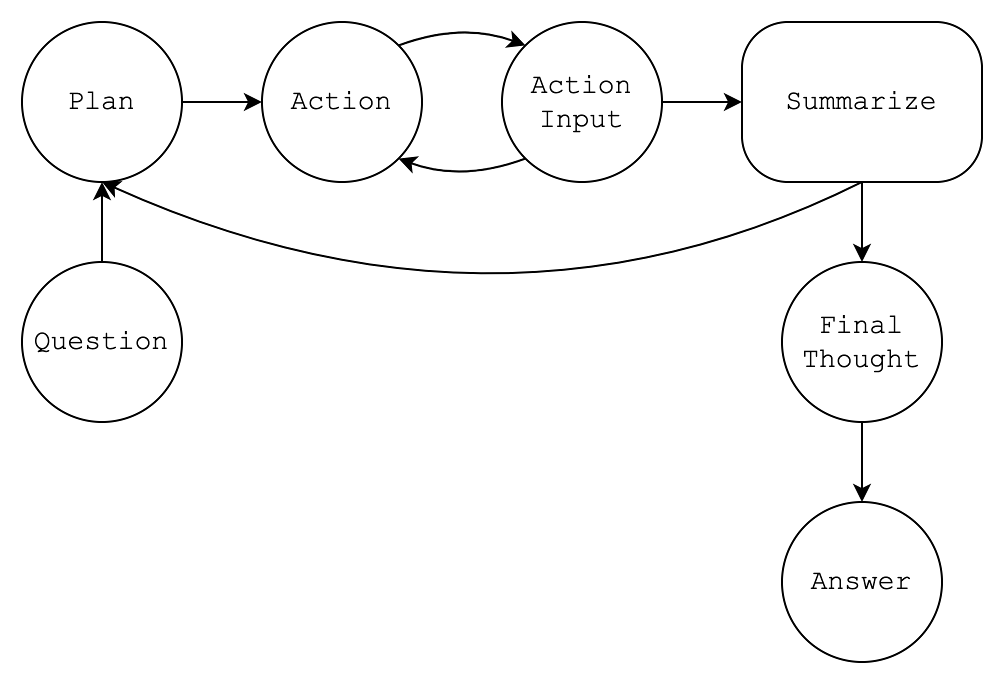}
\caption{State diagram of \sysname{} agent architecture}
\label{fig:pass_arch_diagram}
\end{center}
\end{figure}

In addition to implementing several existing agents (see Appendix \ref{sec:agent_defs}), we also introduce a new agent, referred to as the Plan-Act-Summarize-Solve (\sysname{}) agent. \sysname{} is a hybrid between the entirely sequential ReACT \cite{yao2022react} and the entirely parallel ReWOO \cite{xu2023rewoo} that is designed to address both of their deficiencies. In Figures \ref{fig:pass_spec} and \ref{fig:pass_arch_diagram}, we provide both the specification and finite-state diagram for the \sysname{} agent.

At a high level, \sysname{} operates in a loop that alternates between action execution and summarization. Each iteration of the main loop begins with a planning step (i.e., \texttt{Plan} in Figure \ref{fig:pass_arch_diagram}), where the LLM writes out the series of actions it believes must be executed (i.e., \texttt{Action} and \texttt{Action Input} in Figure \ref{fig:pass_arch_diagram}). Then, the LLM aggregates a set of \textit{independent} actions that can be executed simultaneously in service of the previously written plan. The set of actions is executed, and the results of those actions are summarized by an LLM before being returned to the agent (i.e., \texttt{Summarize} in Figure \ref{fig:pass_arch_diagram}). This loop continues until the agent believes it can solve the problem, at which point it exits the planning loop and writes its final solution (i.e., \texttt{Final~Thought} and \texttt{Answer} in Figure \ref{fig:pass_arch_diagram}).

The summarization step takes as input the original question, the subgoal to be solved, and the result of each action that was executed (in a list format). It then uses the same LLM that powers the agent to produce a summary of the action results with respect to the question and subgoal. Once the summary has been generated, the summarization step must decide whether it is best to return the generated summary or the raw action results to the agent. Letting $x$ be the LLM context thus far, $s$ be the summarization, and $o$ be the original action results concatenated into a list, then the output returned to the \sysname{} agent is
\begin{align*}
    \argmax_{y \in \{ s, o\}} \sum_{i}^{|y|} \log p(y_i | x, y_1, \ldots, y_{i-1}) \big/ \dfrac{(5 + |y|)^{\alpha}}{(5 + 1)^{\alpha}}
\end{align*}
where $p(y | x)$ is computed by the same LLM powering the agent and the score is length-normalized \cite{wu2016google, murray-chiang-2018-correcting}.

In this work, the summarization step is intended to serve two purposes. First, when the summarization is returned, it can consolidate a potentially large amount of information into a concise summary which (ideally) excludes extraneous details that are unhelpful with respect to the original prompt. Second, the summarization step provides the agent with action results that are most in line with the distribution of the underlying model (whether that be the raw action results or the summarization). That is, rather than the agent's LLM incorporating text from arbitrary sources into its generation process, it instead incorporates text produced by an identical LLM (assuming the summarization model is the same as the agent's LLM). In Figure \ref{fig:summarizer_text}, we provide an example of the summarization LLM's input and output.

\begin{figure}[t]
\begin{center}
\begin{tcolorbox}[width=0.97\linewidth,colback=white!90!black]
{\small
\texttt{\textbf{Statements:} Arthur's Magazine (1844-1846) was an American periodical published in Philadelphia in the 19th century. \\
First for Women is a woman's magazine published by Bauer Media Group in the USA.[1] The magazine was started in 1989. \\
\textbf{Context:} Which magazine was started first Arthur's Magazine or First for Women? \\
\textbf{Goal:} I need to search Arthur's Magazine and First for Women, and find which was started first. \\
\textbf{Summary:} Arthur's Magazine started in 1844 and First for Women started in 1989
}
}
\end{tcolorbox}
\caption{Example of inputs (i.e., \texttt{Statements}, \texttt{Context}, and \texttt{Goal}) and output (i.e., \texttt{Summary}) for the summarization step. The summarization LLM was called for the actions \texttt{Search[Arthur's~Magazine]} and \texttt{Search[First~for~Women]}}
\label{fig:summarizer_text}
\end{center}
\end{figure}

The aggregation of a set of actions to take differentiates \sysname{} from ReACT, which executes actions one by one in sequence. Thus, it can be considered the extension of ReACT to partially ordered action spaces. The advantage of this approach is that, like ReWOO, it reduces costly action execution steps that require the agent to suspend the LLM. Unlike ReWOO, however, which cannot leverage the result of an action to influence how it generates subsequent actions (as ReWOO writes out all actions prior to execution), our approach can incorporate feedback from executed actions in its generation process. This can be advantageous when incorrect action calls are possible (e.g., executing a Wikipedia search with the wrong article title).


\section{Experiments}

To demonstrate our framework, we implemented five prompting approaches: 1) Self-consistency alone (Direct), i.e., where the model immediately responds with an answer \cite{wang2022self}, 2) Chain-of-Thought with self-consistency (CoT) \cite{wei2022chain}, 3) ReACT \cite{yao2022react}, 4) ReWOO \cite{xu2023rewoo}, and 5) PASS (our framework, see Section \ref{sec:pass}). In addition, following \cite{yao2022react}, we tested the complimentarity of agent and non-agent-based approaches. For this, questions were first attempted by the non-agent approaches (i.e., Direct / CoT). Then, if no answer was repeated after $k=5$ self-consistency attempts, the agent-based approach was used to produce an answer. In the tables, these results will be indicated with $\texttt{[Non-Agent]}\rightarrow\texttt{[Agent]}$.

We evaluated these approaches on three standard datasets: 1) GSM8K \cite{cobbe2021training}, a mathematical reasoning dataset that tests the ability of a system to solve grade school math word problems, 2) HotpotQA \cite{yang2018hotpotqa}, a multi-hop question-answering dataset that requires reasoning over passages from Wikipedia, and 3) TriviaQA \cite{joshi-etal-2017-triviaqa}, a challenging, question-answering dataset with compositional questions. 

Where possible, we follow the methodology of prior agent-based works in terms of datasets and hyperparameters (see also, Appendix \ref{sec:hparams}), with prompts taken from \cite{xu2023rewoo}. Like \cite{yao2022react,xu2023rewoo}, we do not use the annotated Wikipedia passages for HotpotQA or TriviaQA and instead have the agent choose what terms to search in Wikipedia. To mitigate the cost involved with running large models across a large number of experiments, we follow \cite{xu2023rewoo,shinn2023reflexion,liu2023bolaa,yao2022react} for the larger datasets and select a random subset of questions from their development set for evaluation. For HotpotQA and TriviaQA we select a random subset of 1000 questions from the development set for evaluation, while for GSM8K we keep the full test set for evaluation.



In all experiments, we used Llama-2-70b \cite{touvron2023llama} as the LLM powering our agents. All systems had access to the same tools: 1) \texttt{Calculator} for GSM8K, which executes an input formula and returns a number, 2) \texttt{Search} for HotpotQA and TriviaQA, which returns the first sentences of the Wikipedia page for the entity if it exists, or returns the most similar entities with Wikipedia pages, 3) \texttt{Lookup} for HotpotQA and TriviaQA, which returns the next sentence containing the input string in the last page searched for. 



\section{Results}

{
\setlength{\tabcolsep}{5 pt}
\begin{table}[t]
\centering
\footnotesize
\begin{tabular}{l|ccc}
\toprule
Method & HotpotQA & GSM8K & TriviaQA \\
\midrule
Direct & 29.3 & 15.8 & \bf{74.4} \\
CoT (\citeauthor{wei2022chain}) & \bf{35.0} & 64.6 & 69.4 \\
\midrule
ReACT (\citeauthor{yao2022react}) & 28.7 & \bf{53.8} & 44.1 \\
ReWOO (\citeauthor{xu2023rewoo}) & 27.0 & 51.9 & 52.4 \\
PASS (ours) & \bf{34.7} & 53.2 & \bf{57.8} \\
\midrule
Direct $\rightarrow$ ReACT & 33.1 & 33.3 & 75.3 \\
Direct $\rightarrow$ ReWOO & 31.3 & 32.6 & 75.4 \\
Direct $\rightarrow$ PASS & 33.3 & 33.1 & \bf{75.9} \\
CoT $\rightarrow$ ReACT & 37.8 & 65.3 & 71.4 \\
CoT $\rightarrow$ ReWOO & 37.0 & \bf{65.6} & 71.6 \\
CoT $\rightarrow$ PASS & \bf{38.5} & 64.7 & 72.6 \\
\bottomrule
\end{tabular}
\caption{Exact-match accuracy (\%) results for alternative prompting types across datasets. The best results in all \textbf{three groups} (agent-based, non-agent-based, and hybrid) are bolded}
\label{res:full_results}
\end{table}
}

In Table \ref{res:full_results} we provide the results for all three datasets. We again note that all agents were implemented and evaluated in our framework. While a perfect comparison to prior work is difficult to achieve, as each of \cite{yao2022react,xu2023rewoo,shinn2023reflexion,liu2023bolaa} evaluate on different subsets of each dataset and use different underlying models (e.g., GPT-3.5 \cite{brown2020language}, PaLM \cite{chowdhery2022palm}), our results are consistent in terms of trends with the findings of \cite{xu2023rewoo,yao2022react}. In \cite{xu2023rewoo}, ReWOO was demonstrated to outperform ReACT on TriviaQA (51.8\% versus 47.4\% in their work) and underperform on HotpotQA (30.4\% versus 32.2\%). Lastly, like \cite{yao2022react} and \cite{xu2023rewoo}, we observed chain-of-thought \cite{wei2022chain} and direct response to be more effective than agent-based works.

In addition to individual methods, Table \ref{res:full_results} also shows how well the agent approaches complimented the non-agent-based approaches. As shown in the table, hybrid systems (i.e., non-agent systems that used agents as a fallback prediction) resulted in the top performances across all datasets, with the best results being produced by non-agent methods combined with our \sysname{} agent. This suggests there to be complimentary between these two distinct response paradigms.

\begin{figure}[t]
\begin{center}
\begin{tcolorbox}[width=0.97\linewidth,colback=white!90!black]
{\small
\texttt{\textbf{[Question]} Who was born first, Yanka Dyagileva or Alexander Bashlachev? \\
\textbf{[Thought]} I need to search Yanka Dyagileva and Alexander Bashlachev, find their birth years, then find which was born first. \\
\textbf{[Action]} Search \\
\textbf{[Action Input]} Yanka Dyagileva \\
\textbf{[Action]} Search \\
\textbf{[Action Input]} Alexander Bashlachev \\
\textbf{[Summary]} Yanka Dyagileva was born in 1966 while Alexander Bashlachev was born in 1960. \\ 
\textbf{[Final Thought]} 1960 (Alexander \\ Bashlachev) < 1966 (Yanka Dyagileva), so Alexander Bashlachev was born first. \\ 
\textbf{[Answer]} Alexander Bashlachev
}
}
\end{tcolorbox}
\caption{Example of text output by \sysname{} agent for a question from the HotpotQA dataset}
\label{fig:pass_text}
\end{center}
\end{figure}

{
\setlength{\tabcolsep}{5 pt}
\begin{table}[t]
\centering
\footnotesize
\begin{tabular}{l|ccc}
\toprule
Agent & HotpotQA & GSM8K & TriviaQA \\
\midrule
ReACT (\citeauthor{yao2022react}) & 28.7 & 53.8 & 44.1 \\
ReWOO (\citeauthor{xu2023rewoo}) & 27.0 & 51.9 & 52.4 \\
\midrule
PASS & 34.7 & 53.2 & 57.8 \\
\ \ - No Summarizer & 25.7 & 53.8 & 45.2 \\
\bottomrule
\end{tabular}
\caption{Exact-match accuracy (\%) ablation results for \sysname{} architecture with / without the summarization step}
\label{res:abl_results}
\end{table}
}

Among the agent-based systems, we see that our \sysname{} agent performed the best on HotpotQA and TriviaQA while underperforming as compared to ReACT for GSM8K. We provide an example of \sysname{} output in Figure \ref{fig:pass_text} for the HotpotQA dataset. We suspected the summarization step to have played a key role in the performance of \sysname{}. Thus, we ran an ablation experiment with the \sysname{} agent where the summarization model was removed. In this setting, the results of all actions were instead put into a numbered list and returned to the agent unmodified. 

Table \ref{res:abl_results} shows the results of the ablation experiment. 
In the table, we see that, for HotpotQA and TriviaQA, the use of summarization boosted performance. This could be due to the search and lookup tools often returning very information dense results (i.e., raw Wikipedia text) that contain extraneous information.
In contrast, for GSM8K the summarization LLM negatively impacted results. In GSM8K, the only tool available was a calculator that returned numeric values. We believe that, because there was little to summarize, the LLM could not provide enough value to offset any errors it would make from introduced hallucinations.
These results suggest that summarization can be a very valuable addition to a tool-augmented agent, however, it should only be applied when the raw results of tool calls are likely to derail the agent.



\section{Related Work}



\subsection{LLM-Based Agents}

Various LLM-based agents targeting different tasks have been proposed. Among them, WebAgent \cite{gur2023real} demonstrates language-based agents capable of executing tasks on websites by adhering to natural language commands. Relatedly, the Generative Agents \cite{park2023generative} work aims to simulate believable human behavior, and SayCan \cite{ahn2022can} illustrates the capability to use LLMs in embodied agents. These agents are intended for a particular purpose, and thus are less related to this work than those that can be tailored to solve a wider range of tasks, e.g., ReACT \cite{yao2022react}, Reflexion \cite{shinn2023reflexion}, ReWOO \cite{xu2023rewoo}.

Several open-source projects have also centered on agent creation, e.g., AutoGPT \cite{autogpt2023} and BabyAGI \cite{babyagi2023}. These works have focused on constructing self-sufficient agents that fulfil user requests, which differs from our framework wherein the user defines task-specific individual agents for their specific purposes.

There have been several multi-agent orchestration frameworks that have been recently introduced, e.g., BOLAA \cite{liu2023bolaa}, MetaGPT \cite{hong2023metagpt}, Gentopia \cite{xu-etal-2023-gentopia}, and AutoGen \cite{wu2023autogen}. These frameworks focus primarily on the related problem of orchestrating multiple, simple LLM-based agents to achieve complex goals, where the individual agents themselves are generally no more complex than ReACT or ReWOO. This differs from our focus of designing, implementing, and validating complex individual agents (however, we suspect our work would synergize well with such approaches).

Most conceptually similar to our approach are the works of \cite{langchain2022} and \cite{beurer2023prompting}. LangChain \cite{langchain2022} is a popular framework for designing LLM-based applications that has some support for implementing LLM-based agents. However, their approach does not guarantee conformance to the user's desired behavior, relying almost entirely on prompting to encourage models to follow the expected behavior. The work of \cite{beurer2023prompting} introduced a scripting language for LLMs based on prompting. Their approach did not focus exclusively on LLM-based agents, instead more broadly considering the problem of interleaving LLM outputs into complex, programmatically defined text templates. A key point of differentiation with their approach is that it utilized eager enforcement of constraints (e.g., when used to implement the ReACT agent, their approach halted decoding at each individual state). While this allowed them to capture a much larger range of constraints and behaviors, it also leads to costly reprompting (see \cite{xu2023rewoo}).




\subsection{Constrained Decoding} 

Conceptually related to our work are constrained generation methods that modify the standard beam search decoding procedure at inference time, to incorporate constraints in the output \cite{wiseman-rush-2016-sequence, wuebker-etal-2016-models, anderson-etal-2017-guided,lu2022neurologic,hokamp2017lexically,post2018fast}. While these works focus one level lower than our work in that they go directly into the decoder to modify its outputs, such methods could be used to implement controls within our approach if we were to instead rely on locally hosted models. 

Of particular note, \cite{anderson2017guided} proposes a constrained beam search algorithm that keeps track of constraints via a finite-state machine, and demonstrates its benefits on several image captioning tasks. 
The \textsc{neurologic decoding} line of work \cite{lu2021neurologic, lu2022neurologic} enforces the satisfaction of lexical constraints (specified as any predicate logic formula having word inclusion and exclusion constraints) via adding a penalty term for constraint violation in the beam search decoding algorithm. 
\cite{bastan2023neurostructural} builds on top of \textsc{neurologic decoding} by incorporating structural constraints that capture dependency parsing information. 
Lastly, both the FUDGE \cite{yang2021fudge} and NADO \cite{meng2022controllable} works propose the use of auxillary models that have been trained to recognize constraint-satisfying outputs as a means of controlling a base model (e.g., pretrained LLM) for generation.


\section{Conclusion}

In this work, we introduced a high-level, declarative framework for defining LLM-based agents. We implemented a number of well-known agent types with our framework, and went further to introduce \sysname{}, a new agent architecture that takes advantage of the declarative nature of our framework. Lastly, we compared its performance to other agents across three standard datasets and found it to be strong-performing with complimentary abilities to non-agent-based prompting approaches.


\bibliography{main}
\bibliographystyle{acl_natbib}
\newpage
\appendix

\section{Appendix}

\subsection{Linear Temporal Logic}
\label{sec:ltl_overview}

\begin{figure}[t]
\begin{center}
\includegraphics[width=\linewidth]{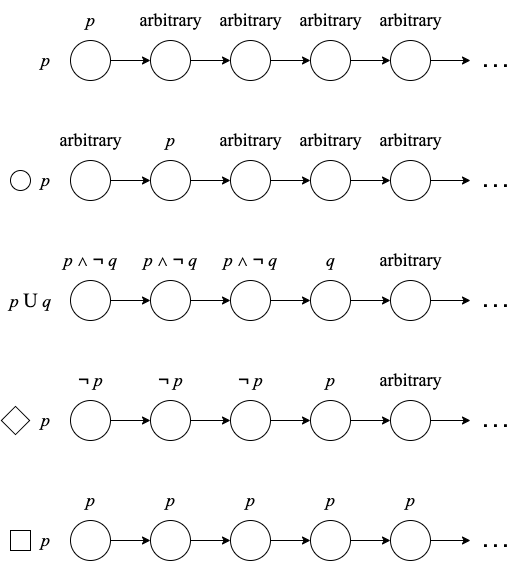}
\end{center}
\caption{Examples of truth assignments over time with various LTL operators}
\label{fig:ltl_ops}
\end{figure}

Here we provide a light overview of linear temporal logic (LTL), which is used in our agent specification framework. LTL is a modal temporal logic originally introduced for formal verification \cite{pnueli1977temporal} that extends propositional logic with the temporal operators $\LtlNext$ (\textit{next}) and $\LtlUntil$ (\textit{until}). The two operators have intuitive definitions, with $\LtlNext$ (\textit{next}) being a unary operator that (informally) means a formula $\varphi$ must hold in the next time step, and $\LtlUntil$ (\textit{until}) being a binary operator that specifies a formula $\varphi_i$ must be true until $\varphi_j$ becomes true. LTL formulas are defined over a set of atomic propositions $\mathcal{P}$ with their syntax given by
\begin{equation*}
\varphi ::= \mathrm{true} \ \ \big| \ \ p \ \ \big| \ \ \neg \varphi \ \ \big| \ \ \varphi_1 \wedge \varphi_2 \ \ \big| \ \ \LtlNext \varphi \ \ \big| \ \ \varphi_1 \LtlUntil \varphi_2
\end{equation*}
where $p \in \mathcal{P}$. An LTL formula is evaluated over an infinite sequence of observations, where each observation is a truth assignment over symbols in $\mathcal{P}$. Letting $\varphi$ be a LTL formula and $\sigma$ be the sequence of observations $\sigma = \left< \sigma_1, \sigma_2, \ldots \right>$, where each $\sigma_i$ can be considered the subset of $\mathcal{P}$ that is true at time $i$, then we write $\sigma \models \varphi$ (satisfies) when
\begin{alignat*}{4}
&\sigma \quad &&\models \quad \mathrm{true} \quad && && \\
&\sigma \quad &&\models \quad p \quad && \textrm{iff} \quad p \in \sigma_0 \\
&\sigma \quad &&\models \quad \neg \varphi \quad && \textrm{iff} \quad \sigma \not\models \varphi \quad \quad && \\
&\sigma \quad &&\models \quad \varphi_1 \wedge \varphi_2 \quad && \textrm{iff} \quad \sigma \models \varphi_1 \  \ \textrm{and} \ \ \sigma \models \varphi_2 \\
&\sigma \quad &&\models \quad \LtlNext \varphi \quad && \textrm{iff} \quad \sigma[1\ldots] \models \varphi \\
&\sigma \quad &&\models \quad \varphi_1 \LtlUntil \varphi_2 \quad && \textrm{iff} \quad \exists j \geq 0 . \ \ \sigma[j\ldots] \models \varphi_2 && \\
& && && \textrm{and} \ \ \sigma[i\ldots] \models \varphi_1 \\
& && && \textrm{for all} \ \  0 \leq i < j
\end{alignat*}
where $\sigma[i\ldots] = \left< \sigma_i, \ldots \right>$ is the remaining sequence of observations following time step $i$. From the operators listed above, we can define additional propositional logic operators $\vee$ (\textit{disjunction}), and $\rightarrow$ (\textit{implication}) as well as temporal operators $\LtlEventually$ (\textit{eventually}) and $\LtlAlways$ (\textit{always})
\begin{alignat*}{2}
&\varphi_1 \vee \varphi_2 \quad &&\coloneqq \quad \neg (\neg \varphi_1 \wedge \neg \varphi_2) \\
&\varphi_1 \rightarrow \varphi_2 \quad &&\coloneqq \quad \neg \varphi_1 \vee \varphi_2 \\
&\LtlEventually \varphi \quad &&\coloneqq \quad \mathrm{true} \LtlUntil \varphi \\
&\LtlAlways \varphi \quad &&\coloneqq \quad \neg \LtlEventually \neg \varphi
\end{alignat*}
In this work, we treat \textit{next} as an n-ary operator, which can be considered simply a chained sequence of \textit{next} operators, i.e., $\LtlNext (\varphi_1, \varphi_2, \varphi_3, \ldots)$ with the straightforward informal interpretation of ``$\varphi_1$ then $\varphi_2$ then $\varphi_3$'', etc.. In Figure \ref{fig:ltl_ops}, we provide a graphical depiction of the truth assignments over time for the above temporal operators. 

In this work, we found that only allowing formulas to be those containing atomic propositions $p$, as well as operators $\rightarrow$, $\LtlNext$, $\LtlAlways$, and $\LtlUntil$ (i.e., we do \textit{not} allow formulas to include $\neg$, $\wedge$, etc.) was sufficient to represent the range of existing agent architectures. We leave extending the set of operators (e.g., to include $\LtlEventually$, $\wedge$, etc.) to future work. For more details regarding LTL and its numerous applications, we direct the interested reader to \cite{baier2008principles}.

\subsection{Hyperparameters and Hardware}
\label{sec:hparams}
There were not many hyperparameters to our approach beyond those listed in the experiments section, as our LLMs were only used for inference. The decoding strategy for our LLMs was set to greedy in all cases except for approaches which used self-consistency (i.e., Direct, CoT). For self-consistency, the temperature for sampling was set to 0.7 (as in \cite{yao2022react}) and $k=5$ samples were drawn.

\subsection{Agent Definitions}
\label{sec:agent_defs}

\begin{figure}
\begin{center}
\begin{tcolorbox}[width=0.97\columnwidth,colback=white!90!black]
{\small
\texttt{(define react-agent \\
\null\quad(:states \\
\null\quad\quad(Ques (:text "[Question]")) \\
\null\quad\quad(Tht (:text "[Thought]")) \\
\null\quad\quad(Act (:text "[Action]")) \\
\null\quad\quad(Act-Inp (:text "[Action Input]")) \\
\null\quad\quad(Obs (:text "[Observation]") \\
\null\quad\quad\quad\quad \ (:flags :env-input)) \\
\null\quad\quad(Final-Tht (:text "[Final Thought]")) \\
\null\quad\quad(Ans (:text "[Answer]"))) \\
\null\quad \\
\null\quad(:behavior \\
\null\quad\quad(next \\
\null\quad\quad\quad Ques \\
\null\quad\quad\quad(until \\
\null\quad\quad\quad\quad(next Tht Act Act-Inp Obs) \\
\null\quad\quad\quad\quad Final-Tht) \\
\null\quad\quad\quad Ans)))
}
}
\end{tcolorbox}
\end{center}
\caption{Specification for ReACT agent \cite{yao2022react}}
\label{fig:app_react_spec}
\end{figure}

\begin{figure}
\begin{center}
\begin{tcolorbox}[width=0.97\columnwidth,colback=white!90!black]
{\small
\texttt{(define rewoo-agent \\
\null\quad(:states \\
\null\quad\quad(Ques (:text "[Question]")) \\
\null\quad\quad(Plan (:text "[Plan]")) \\
\null\quad\quad(Act-Lbl (:text "[Action Label]")) \\
\null\quad\quad(Act (:text "[Action]")) \\
\null\quad\quad(Act-Inp (:text "[Action Input]")) \\
\null\quad\quad(Solver (:text "[Answer]") \\
\null\quad\quad\quad\quad\quad\quad (:flags :env-input))) \\
\null\quad \\
\null\quad(:behavior \\
\null\quad\quad(next \\
\null\quad\quad\quad Ques \\
\null\quad\quad\quad(until \\
\null\quad\quad\quad\quad(next Plan Act-Lbl Act Act-Inp) \\
\null\quad\quad\quad\quad Solver))))
}
}
\end{tcolorbox}
\end{center}
\caption{Specification for ReWOO agent \cite{xu2023rewoo}}
\label{fig:app_rewoo_spec}
\end{figure}

\begin{figure}
\begin{center}
\begin{tcolorbox}[width=\columnwidth,colback=white!90!black]
{\small
\texttt{(define reflexion-agent \\
\null\quad(:states \\
\null\quad\quad(Ques (:text "[Question]")) \\
\null\quad\quad(Tht (:text "[Thought]")) \\
\null\quad\quad(Act (:text "[Action]")) \\
\null\quad\quad(Act-Inp (:text "[Action Input]")) \\
\null\quad\quad(Obs (:text "[Observation]") \\
\null\quad\quad\quad\quad \ (:flags :env-input)) \\
\null\quad\quad(Final-Tht (:text "[Final Thought]")) \\
\null\quad\quad(Prop-Ans (:text "[Proposed Answer]")) \\
\null\quad\quad(Eval (:text "[Evaluation]") \\
\null\quad\quad\quad\quad\quad (:flags :env-input)) \\
\null\quad\quad(Ref (:text "[Reflection]")) \\
\null\quad\quad(Ans (:text "[Answer]"))) \\
\null\quad \\
\null\quad(:behavior \\
\null\quad\quad(next \\
\null\quad\quad\quad Ques \\
\null\quad\quad\quad(until \\
\null\quad\quad\quad\quad(next \\
\null\quad\quad\quad\quad\quad(until \\
\null\quad\quad\quad\quad\quad\quad(next Tht Act Act-Inp Obs) \\
\null\quad\quad\quad\quad\quad\quad Final-Tht) \\
\null\quad\quad\quad\quad\quad Prop-Ans \\
\null\quad\quad\quad\quad\quad Eval \\
\null\quad\quad\quad\quad\quad Ref) \\
\null\quad\quad\quad\quad Ans))))
}
}
\end{tcolorbox}
\end{center}
\caption{Specification for Reflexion agent \cite{shinn2023reflexion}}
\label{fig:app_reflexion_spec}
\end{figure}

\begin{figure}
\begin{center}
\begin{tcolorbox}[width=0.97\columnwidth,colback=white!90!black]
{\small
\texttt{(define cot-agent \\
\null\quad(:states \\
\null\quad\quad(Ques (:text "[Question]")) \\
\null\quad\quad(Tht (:text "[Thought]")) \\
\null\quad\quad(Ans (:text "[Answer]"))) \\
\null\quad \\
\null\quad(:behavior \\
\null\quad\quad(next Ques Tht Ans)))
}
}
\end{tcolorbox}
\end{center}
\caption{Specification for Chain-of-thought agent \cite{wei2022chain}}
\label{fig:app_cot_spec}
\end{figure}

\begin{figure}
\begin{center}
\begin{tcolorbox}[width=0.97\columnwidth,colback=white!90!black]
{\small
\texttt{(define direct-agent \\
\null\quad(:states \\
\null\quad\quad(Ques (:text "[Question]")) \\
\null\quad\quad(Ans (:text "[Answer]"))) \\
\null\quad \\
\null\quad(:behavior \\
\null\quad\quad(next Ques Ans)))
}
}
\end{tcolorbox}
\end{center}
\caption{Specification for direct response agent (i.e., outputs only answer)}
\label{fig:app_direct_spec}
\end{figure}

\end{document}